\pdfoutput=1

\documentclass[11pt]{article}

\usepackage{ACL2023}

\usepackage{times}
\usepackage{latexsym}
\usepackage{graphicx}
\usepackage{multirow}

\usepackage[T1]{fontenc}

\usepackage[utf8]{inputenc}

\usepackage{microtype}

\usepackage{inconsolata}

\title{Unsupervised Domain Adaptation using Lexical Transformations and Label Injection for Twitter Data}

\author{Akshat Gupta, Xiaomo Liu, Sameena Shah \\
  J.P.Morgan AI Research \\
  \texttt{ \{akshat.x.gupta, xiaomo.liu, sameena.shah\}@jpmorgan.com }}

\begin{document}
\maketitle
\begin{abstract}
Domain adaptation is an important and widely studied problem in natural language processing. A large body of literature tries to solve this problem by adapting models trained on the source domain to the target domain. In this paper, we instead solve this problem from a dataset perspective. We modify the source domain dataset with simple lexical transformations to reduce the domain shift between the source dataset distribution and the target dataset distribution. We find that models trained on the transformed source domain dataset performs significantly better than zero-shot models. Using our proposed transformations to convert standard English to tweets, we reach an unsupervised part-of-speech (POS) tagging accuracy of 92.14\% (from 81.54\% zero shot accuracy), which is only slightly below the supervised performance of 94.45\%. We also use our proposed transformations to synthetically generate tweets and augment the Twitter dataset to achieve state-of-the-art performance for POS tagging.
\end{abstract}

\section{Introduction}
In a typical machine learning setting, training, development and test sets are usually carved out of the same data collection effort. In doing this, we caveat our models with an implicit assumption - the deployment dataset should belong to the same distribution as the training dataset. This is rarely the case and we see significant drops in performance when the model is deployed. The mismatch between the deployment data distribution, or \textit{target domain}, and the training data distribution, or \textit{source domain}, is known as domain shift \citep{ramponi2020neural, ruder2018strong} and the process of adapting to target domain distributions is known as domain adaptation \cite{blitzer2006domain}.

The most widely studied domain adaptation methods are model-centric methods \cite{ramponi2020neural}, where parts of the model, including the feature space, the loss function or even the structure of the model are altered \citep{blitzer2006domain, pan2010cross, ganin2016domain, marz2019domain}. Data-centric methods \cite{ramponi2020neural} usually involve some form of bootstrapping and pseudo-labelling of the target domain data \citep{abney2007semisupervised, cui2019self, ruder2018strong, gupta2021unsupervised}. A popular data-centric domain adaptation method is data selection, which is an intermediate training step that aims to select a subset of data that is closest to the target domain \cite{moore2010intelligent, axelrod2011domain, aharoni2020unsupervised, iter2021complementarity}. We refer the reader to domain adaptation surveys in natural language processing for a detailed overview \citep{ramponi2020neural, chu2018survey, jiang2013literature, margolis2011literature}.

\begin{figure}
 \centering
 \scalebox{0.15}{
    \centering
     \includegraphics{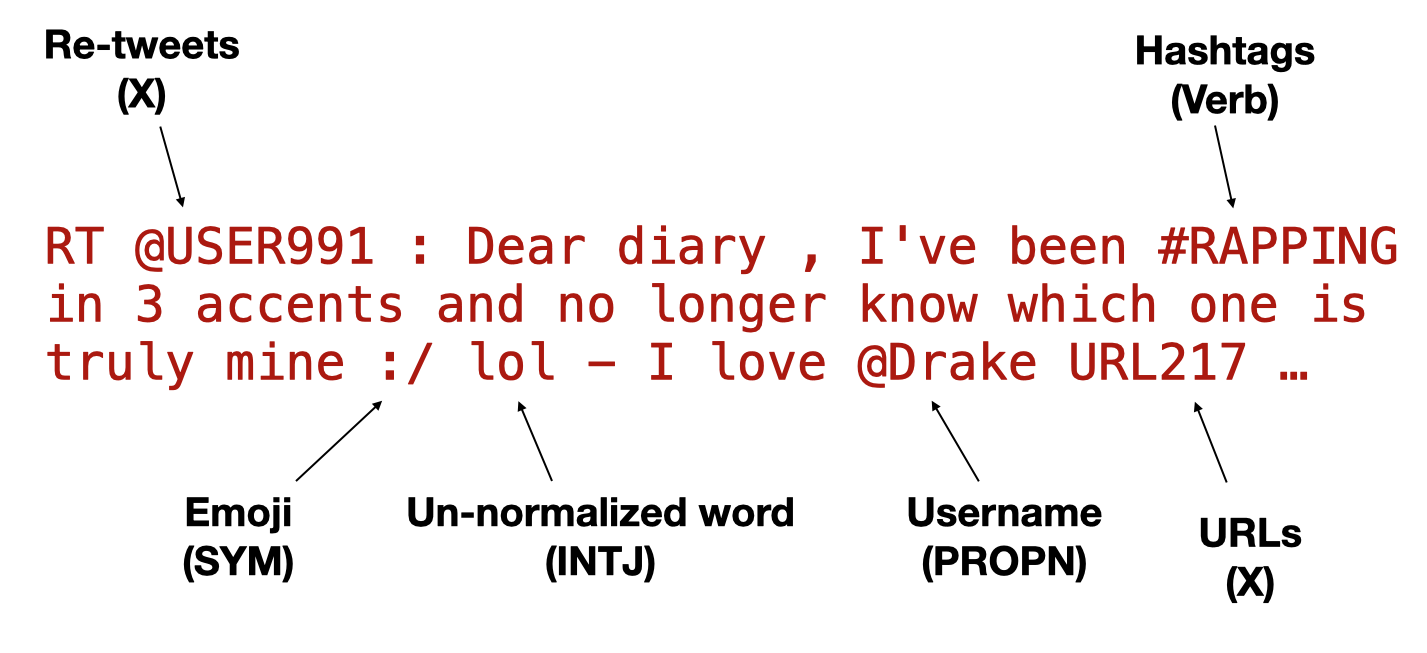}}
     \caption{The Anatomy of a Tweet - This figure shows lexical features of a tweet like hashtags, user-mentions, emojis, re-tweets.}
     \label{fig:anatomy}
\end{figure}

 To the best of our knowledge, none of the works we encounter in literature address the fundamental reason behind the need for domain adaptation - \textit{domain shift}. If we are able to transform the source domain dataset such that the domain mismatch between the source domain and the target domain is reduced, while being able to exploit the annotations of the source domain corpus, then the models trained on such a transformed source domain data will naturally perform better on the target domain. This is the main motivation behind our work. All model-centric and data-centric domain adaptation methods can be applied on top of our proposed method and are complementary to it.
 
 In this paper, we transform the source domain dataset to resemble the target domain dataset more closely through a series of transformations. In our case, the source domain consists of standard English sentences and the target domain consists of tweets. Through these transformations, we are able to improve the zero-shot POS tagging accuracy by 10.39\% when averaged over five different BERT models. Also, when we combine the transformed data to augment the original target dataset, we achieve state-of-the-art POS tagging performance on the target dataset. 

  \begin{figure}
 \centering
 \scalebox{0.15}{
    \centering
     \includegraphics{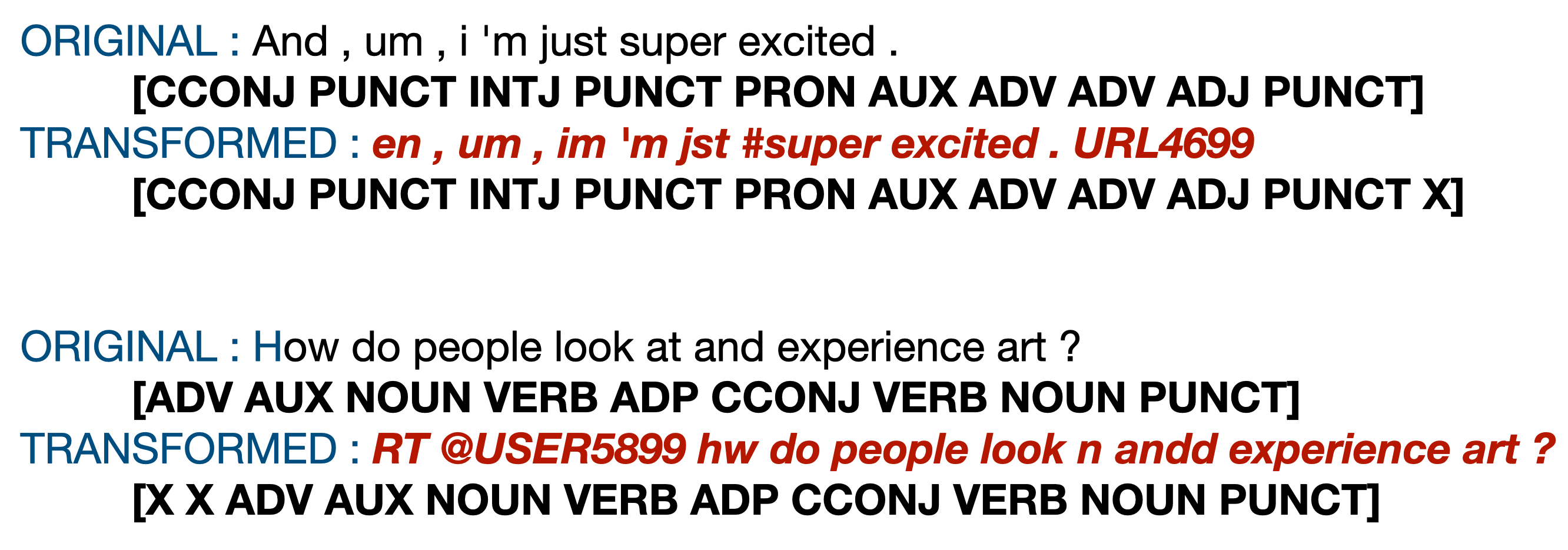}}
     \caption{Examples of original sentences from GUM dataset and how they look like after different Lexical Transformations and Label Injections. Note the POS labels injected post transformations.}
     \label{fig:example}
\end{figure}

\section{Lexical Transformations and Label Injections}
 Standard English sentences and Tweets have both semantic and lexical differences. Tweets are more likely to be subjective and polarized (appendix \ref{appendix:dataset}). On the other hand, tweets also contain unique lexical features like acronyms, emojis, user mentions, retweets, hashtags, as shown in Figure \ref{fig:anatomy}, and can be used as different parts of speech (Table \ref{table:features}, appendix \ref{appendix:lexical features}). In this paper, we focus on converting standard English sentences into tweets by making lexical transformations and injecting labels wherever required. Example transformations are shown in Figure \ref{fig:example}.
 
 Lexcial transformations add target domain-specific lexical features to the source domain dataset such that these properties are `distributionally' conserved. For example, when our target domain is Twitter, we expect Tweets to contain emojis. We can measure the distributional presence of emojis in tweets, like the percentage of tweets that on average contain emojis or how they are distributed within the sentence, i.e. if they are more likely to occur in the beginning, middle, or end of a sentence. In lexical transformations, we add these distributional properties to the source domain sentences. Since we are adding these features to an annotated dataset, we also inject the label of the lexical feature wherever required. The process is discussed in detail in section \ref{sec:experiments}. The resulting sentences are almost indistinguishable from Tweets, as can be seen in Figure \ref{fig:example}. It is not trivial to inject these lexical features into the standard English sentences as the same feature can correspond to multiple parts of speech, as shown in Table \ref{table:features}.
 
 
\begin{table}
\centering
 \scalebox{0.8}{

    \begin{tabular}{l|l|l|l}
    \textbf{Dataset} & \textbf{Split} & \textbf{Sentences} & \textbf{Tokens}\\
    \hline
    GUM & Train & 6,917 & 124,923\\
    TBv2 & Train & 1,639 & 24,753\\
     & Test & 1,201 & 19,911\\
    
    \end{tabular}
    
    }
\caption{\label{table:dataset}
Table showing the dataset statistics for GUM and Tweebank datasets (TBv2). In this paper, all models are tested on the test set of TBv2, which is our target domain set. For compelete statistics, refer to \ref{appendix:dataset}.
}
\end{table}

\begin{table*}
\centering
 \scalebox{0.85}{

    \begin{tabular}{l|l|l|l|l|l|l}
    \textbf{Method} & \textbf{POS}  & \textbf{BERT-base} & \textbf{RoBERTa-base} & \textbf{BERT-large} & \textbf{RoBERTa-Large} & \textbf{BERTweet}\\
    \hline
    Zero Shot & - & 79.74 (0.26) & 80.87 (0.19) & 81.52 (0.23) &  81.83 (0.23) & 80.73 (0.25)\\
    Emoji Injection & SYM & 80.12 (0.19) & 80.85 (0.43) & 81.08 (0.68) & 81.77 (0.51) & 81.59 (0.60) \\
    ILN & - & 80.66 (0.23) & 79.43 (0.18) & 81.33 (0.13) &  80.95 (0.45) & 80.46 (0.34) \\
    @,\#ht & PROPN & 81.89 (0.06) & 80.87 (0.08) & 82.28 (0.04) & 81.92 (0.19) & 82.15 (0.14)\\
    RT,\#ht,url & X & 89.33 (0.08) & 88.09 (0.12) & 89.76 (0.09) & 88.95 (0.21) & 88.89 (0.17) \\
    \end{tabular} }
    
\caption{\label{table:single_exp}
This table shows the performance boost for unsupervised domain adaptation from standard English to Tweets when the four different lexical transformations are used individually.
}
\end{table*}

\section{Datasets}
In this paper, we work with two annotated POS tagging datasets. For standard English, we use the GUM (Georgetown University Multilayer Corpus) dataset \cite{zeldes2017gum}. For Twitter data, we use Tweebank (TBv2) \cite{liu2018parsing} dataset. We choose these two datasets because they are both labelled using the universal dependencies \cite{nivre2016universal} framework, thus each of the datasets have identical 17 POS tags. The dataset statistics are shown in Table \ref{table:dataset}. 

The GUM dataset acts as our source domain dataset and is about 5 times larger than TBv2, which is our target domain dataset. GUM dataset is made up of articles and interviews from Wikinews, instructional articles from wikiHow and travel guides from Wikivoyage \cite{zeldes2017gum}. The GUM dataset contains longer sentences compared to the Tweebank dataset. The Tweebank dataset gets higher average polarity and subjectivity scores when compared to the GUM dataset. The experiments analysing dataset properties are shared in appendix \ref{appendix:dataset}.

\section{Experiments}\label{sec:experiments}
In this section, we present four  different types of Lexical Transformations and corresponding label injection methods for Twitter as target domain. All transformations are performed on the GUM train-split (the standard English dataset). Models trained on the transformed dataset are tested on the TBv2 test set (the Twitter dataset). All experiments shown in this paper report accuracy scores on TBv2 test set, in accordance with previous works \citep{owoputi2013improved, meftah2019joint, meftah2020multi, nguyen2020bertweet}. Each experiment is repeated five times and the mean score is reported with standard deviations reported in brackets.

All experiments in this paper are done using the Huggingface implementations of different BERT models. We use five different BERT models, the original BERT-base-uncased and BERT-large-uncased \cite{devlin2018bert} models, the RoBERTa-base and RoBERTa-large models \cite{liu2019roberta} and the BERTweet model \cite{nguyen2020bertweet}.

\subsection{Zero-Shot Experiments}
We begin by training the model on the original GUM train-split and testing it on the TBv2 dataset. This experiment sets our baseline for unsupervised domain adaptation as it represents zero-shot application of a model trained on standard English, and then applied to tweets. The results are shown as the Zero Shot results of Table \ref{table:single_exp}. 

\subsection{Emoji Injections}
Social media text is filled with emojis and emoticons. In this paper, we refer to both as Emojis. To convert standard English sentences to Tweets, we inject emojis into standard English sentences. Emojis belong to the `SYM:symbol' class in the universal dependencies framework, which is inserted as the label for the injected emoji in the source domain dataset.

To place an emoji within a standard English sentence, we first randomly select an emoji from a pre-decided list of emojis. Then we place the emoji inside a sentence according to a Gaussian distribution which is fit to the location of occurrence of emojis in a tweet. We randomly add emojis to 25\% of the sentences in the GUM dataset. The different experiments done to reach the above methodology for emoji injection are described in appendix \ref{appendix:emoji}. The results for emoji injection are shown in Table \ref{table:single_exp}.

\begin{table*}
\centering
 \scalebox{0.85}{

    \begin{tabular}{l|l|l|l|l|l|l}
   \textbf{Type} & \textbf{Train Dataset} & \textbf{BERT-base} & \textbf{RoBERTa-base} & \textbf{BERT-large} & \textbf{RoBERTa-Large} & \textbf{BERTweet}\\
    \hline
    Unsupervised & GUM & 79.74 (0.26) & 80.87 (0.19) & 81.52 (0.23) &  81.83 (0.23) & 80.73 (0.25)\\
     & \textbf{GUM-T (UDA)} &  \textbf{91.82 (0.07)} & \textbf{90.85 (0.08)} & \textbf{92.14 (0.12)} & \textbf{  90.86 (0.61)} & \textbf{90.99 (0.24)} \\
     \hline
    Supervised & TBv2 & 93.88 (0.05) & 93.00 (0.03) & 94.45 (0.04) &  93.85 (0.08) & 93.85 (0.09)\\
    & TBv2 + GUM & 94.31 (0.06) & 94.16 (0.06) & 94.51 (0.05) & 94.61 (0.08) & 94.71 (0.08)\\
    & \textbf{TBv2 + GUM-T} & \textbf{94.81 (0.02)} & \textbf{94.84 (0.06)} & \textbf{95.01 (0.05)} & \textbf{95.04 (0.04)} & \textbf{95.21 (0.03)} \\
    \end{tabular} }
    
\caption{\label{table:combination_exp}
This table shows the final Unsupervised Domain Adaptation performance using our proposed data transformations. All these models are test on the TBv2 test set and trained on the datasets described above. We combine the transformed data with the original dataset to achieve state-of-the-art results on the Tweebank test set.
}
\end{table*}

\subsection{Inverse Lexical Normalization}
Lexical normalization is a common task where non-standard English tokens are corrected to standard English \cite{han2013lexical}. This includes expanding acronyms like \textit{wru -> where are you} and correcting spelling errors. In this paper, we convert standard English to its lexically un-normalized version. We call this process Inverse Lexical Normalization (ILN). To do so, we use a lexical normalization dataset \cite{baldwin2015shared} as a dictionary lookup and create a mapping between lexically correct words and their un-normalized version. For example, \textit{you} is written in various different ways including \textit{u}, \textit{uuuu}, \textit{youuuu}. We randomly replace the correct tokens with their un-normalized versions 75\% of the times. The ablation experiments for this lexical transformation are shown in \ref{appendix:iln}. The POS tag of the original word is retained in the transformation. BERT-base observes maximum improvement with ILN (Table \ref{table:single_exp}). 


\subsection{Converting PROPN to User-Mentions and Hashtags}

Another distinguishing lexical features of Tweets is the use of user-mentions and hashtags. In this transformation, we randomly pick existing proper nouns in the GUM dataset and convert them into user-mentions or hashtags by adding an '@' or '\#' symbol in front of the token, with a probability of 50\% and 20\% respectively. The existing proper noun labels are kept for the converted tokens. The ablations for this transformation can be found in appendix \ref{appendix:propn}. We see consistent improvements with this transformation for all models except RoBERTa models (Table \ref{table:single_exp}). 

\subsection{Injecting ReTweets, URLS, user-mentions and hashtags as X}
The `X' part of speech tag or the \textit{other} category in the universal dependency framework \cite{nivre2016universal} is defined as - "\textit{The tag X is used for words that for some reason cannot be assigned a real POS category. It should be used very restrictively}". While the `X' POS tag is used sparingly in standard English, a large number of tokens in tweets fall into this category. In this transformation, we insert re-tweets (at the beginning of sentences), urls (usually at the back of the sentences) and hashtags (randomly sampled from a Gaussian calculated from tweets). Re-tweets are added in 30\% of the sentences, URL's are added in 60\% of the sentences and hashtags are added in 10\% of the sentences. The ablations can be found in appendix \ref{appendix:X}. The label `X' is added with these lexical transformations. 


We see massive improvements across the board by adding this lexical transformation. This is because the `X' POS tag, which is probably the most under-utilized tag when dealing with standard English, becomes vital when dealing with tweets. All Re-tweets, URL's and many hashtags and user mentions fall under this category.

\section{Results}
We now combine all transformations together, as shown in Table \ref{table:combination_exp}. The first section in Table \ref{table:combination_exp} represents our unsupervised domain adaptation results. The first row in Table \ref{table:combination_exp} shows models trained on the original GUM dataset (standard English) and tested on TBv2 test set, representing zero-shot domain transfer results. The GUM-T dataset represents the transformed dataset containing all the previously described transformations. Models trained on the GUM-T dataset represent our unsupervised domain adaptation performance, which improves on the zero-shot POS tagging accuracy by 10.39\%, without ever seeing a single tweet (when averaged over all five models). The class-wise F1 improvements for different POS tags are shown in Table \ref{table:classwise}. BERT-base witnesses the maximum gain from our transformations (12.08\%) and performs better than RoBERTa-large and BERTweet. 

The second section in Table \ref{table:combination_exp} contains supervised experiments where the training dataset contains tweets. We check the efficacy of our proposed transformations as a synthetic data generation process. We first augment the TBv2 dataset with the  original GUM dataset and compare it with the improvements we get when TBv2 is combined with GUM-T. We see that the combination of TBv2 and GUM-T datasets outperforms all supervised models and gives 1.6 to 8 times larger performance boost over augmenting with the original GUM dataset. The TBv2 + GUM-T combination reaches (a saturated) state-of-the-art maxima for POS tagging on the TBv2 dataset, as shown in Table \ref{table:sota}.

\begin{table}
\centering
 \scalebox{0.8}{

    \begin{tabular}{ll}
    \hline
    \textbf{System} & \textbf{POS Accuracy} \\
    \hline
    \cite{owoputi2013improved} & 94.6\\
    \cite{meftah2019joint} & 94.95\\ 
    \cite{nguyen2020bertweet} & 95.2\\
    \textbf{BERTweet [TBv2 + GUM-T] (ours)} & \textbf{95.21 (0.03)} \\
    
    \hline
    \end{tabular}
    
    }
\caption{\label{table:sota}
Table showing the dataset statistics for GUM and Tweebank datasets (TBv2). In this paper, all models are tested on the test set of TBv2, which is our target domain set.
}
\end{table}

\section{Conclusion}
A lot of focus in literature has been given to converting noisy social media text to standard English. In our work, we convert standard English into noisy social media-like text using simple lexical transformations and show that it can be used as an effective unsupervised domain adaptation and data augmentation method. The fundamental idea behind our work is to reduce domain shift by transforming the source domain into the target domain. We present experiments for these transformations between standard English and Twitter domain and find an average accuracy boost for POS tagging of 10.39\% across 5 different BERT models, without ever using a single tweet for supervised training.

\section*{Acknowledgements}
This paper was prepared for informational purposes in part by the Artificial Intelligence Research Group of JPMorgan Chase \& Co and its affiliates (“J.P. Morgan”) and is not a product of the Research Department of J.P. Morgan.  J.P. Morgan makes no representation and warranty whatsoever and disclaims all liability, for the completeness, accuracy, or reliability of the information contained herein.  This document is not intended as investment research or investment advice, or a recommendation, offer, or solicitation for the purchase or sale of any security, financial instrument, financial product, or service, or to be used in any way for evaluating the merits of participating in any transaction, and shall not constitute a solicitation under any jurisdiction or to any person if such solicitation under such jurisdiction or to such person would be unlawful. 

© 2022 JPMorgan Chase \& Co. All rights reserved.

\section{Limitations}
In this paper, we focus on lexical transformations between source domain and target domain to reduce the domain shift between them. To do this, we identify unique lexical features in the target domain and place them in the source domain so that the transformed domain is distributionally similar to the target domain. But there are also semantic differences between the two domains in terms of content, domain-specific jargon, and other nuances. This work does not take into account those transformations. Also, we use Twitter as the target domain for our work. While the general principles of our work are applicable to any source-target domain pairs, the transformations discussed in this work cater broadly to social media text, and specifically to Twitter data. The generalizability to other target domains has not been tested in this paper and remains a topic of further investigation.

In this paper, we work with a POS tagging dataset. POS tagging is a token level task where we classify each token as belonging to a certain category. We feel that because POS tagging is dependent on each token in the sentence, domain transfer affects this task most adversely. Sequence classification tasks like sentiment analysis that only require a high level representation of the entire sentence to make classification decisions might witness different levels of improvement. The current method needs to be tested for other task types, including sequence classification tasks like sentiment analysis, or generative tasks like question answering and text summarization. This was beyond the scope of a short paper.

\bibliography{anthology,custom}
\bibliographystyle{acl_natbib}

\newpage
\appendix

\section{Appendix}

\begin{table}
\centering
\begin{tabular}{l|l|l|l}
\textbf{Dataset} & \textbf{Split} & \textbf{Sentences} & \textbf{Tokens}\\
\hline
GUM & Train & 6,917 & 124,923\\
 & Dev & 1,117 & 19,654\\
 & Test & 1,096 & 19,911\\
TBv2 & Train & 1,639 & 24,753\\
 & Dev & 7,10 & 11,759\\
 & Test & 1,201 & 19,911\\

\end{tabular}
\caption{\label{table:dataset_complete}
Table showing the complete dataset statistics for GUM and Tweebank datasets (TBv2).
}
\end{table}

\subsection{Dataset}\label{appendix:dataset}
In this paper, we work with two part-of-speech (POS) tagging datasets. The GUM dataset \cite{zeldes2017gum}, which is made up of standard English sentences from different wiki-sources like wikiNews, wikiHow etc., and the Tweebankv2 (TBv2) dataset \cite{liu2018parsing}, which consists of tweets. The GUM dataset acts as our source domain dataset, while TBv2 acts as our target domain dataset. 

\begin{figure}
 \centering
 \scalebox{0.5}{
    \centering
     \includegraphics{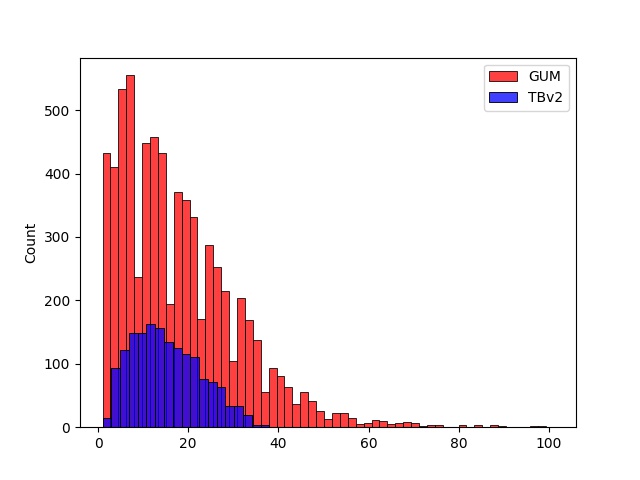}}
     \caption{Sentence length distribution between GUM and Tweebankv2 (TBv2) dataset. We see that GUM has more longer and shorter sentences compared to TBv2.}
     \label{fig:token_dist}
\end{figure}

The number of sentences and the number of tokens in each dataset are given in Table \ref{table:dataset_complete}. Figure \ref{fig:token_dist} shows the sentence length distribution between the GUM and the TBv2 dataset. We see that the GUM dataset contains longer sentences when compared to the TBv2 dataset. The mean tokens per sentence for GUM is 18.06 (std = 13.3) whereas the mean tokens per sentence for the TBv2 dataset is 15.10 (std = 7.74). This shows us that TBv2 not only has shorter sentences, but their spread is also shorter.

We measure average subjectivity and polarity scores for the two datasets to indicate semantic differences. We find higher average subjectivity and polarity scores for the TBv2 dataset compared to the GUM dataset. To measure these, we use the spaCY textblob \footnote{ \url{https://spacy.io/universe/project/spacy-textblob} } library to calculate subjectivity and polarity scores. Polarity is scored between -1 and 1 indicating the sentiment expressed in the sentence. We take the absolute value of the polarity scores since we consider both positive and negative sentiment since we are interested in the presence and absence of polarity in tweets. The mean polarity score for the TBv2 dataset was 0.23 compared to 0.13 for the GUM dataset. Subjectivity is scored between 0 and 1, with 0.0 being very objective and 1.0 being very subjective. TBv2 had a mean subjectivity score of 0.36 compared to 0.27 for the GUM dataset.

\subsection{Lexical Features}\label{appendix:lexical features}
Some of the lexical features specific to tweets that we are concerned with in this paper are - emojis, re-tweets, user-mentions, hashtags, URL's and un-normalized tokens. It is not trivial to inject these into the standard English sentences as same lexical feature can correspond to multiple parts of speech. This can also be seen in Figure \ref{fig:anatomy}, where user-mentions are used both for the category 'X' as well as proper nouns. A more detailed description of the different lexical features and the corresponding parts of speech the features can take can be seen in Table \ref{table:features}. Lexical features like user-mentions can take two parts of speech, where hashtags and un-normalized words can essentially be any part of speech.

\begin{table}
\centering
\begin{tabular}{ll}
\hline
\textbf{Lexical Features} & \textbf{Associated POS tags} \\
\hline
Emoji & SYM\\
Re-Tweets & X\\ 
URLs & X \\
User Mentions & X, PROPN\\
Hashtags & X, VERB, PROPN ...\\
Un-normalized words & VERB, INTJ, ADP ...\\

\hline
\end{tabular}
\caption{\label{table:features}
Table showing the different lexical features seen in tweets and the corresponding part of speech tags the features can take.
}
\end{table}

\subsection{Emoji Injections Ablation}\label{appendix:emoji}
Emoji Injection is a lexical transformation where we insert emojis in standard English sentences such that the distributional properties of the transformed text resemble a Twitter dataset. Lexical emoji injection is done in two steps:

\begin{itemize}
    \item Emoji Selection - Sample an emoji from a pre-selected list of emojis
    \item Emoji Placement - Select a location in the standard English sentence to place the selected emoji
\end{itemize}

Both these steps can be done randomly or based on a particular distribution. The selection step can be done by selecting an emoji based on the distribution of its occurrence in Twitter feeds. Although in this paper, in the emoji selection step, we select an emoji randomly from a pre-decided list of emojis. 

Similarly, the emoji placement step can be done in two ways. The selected emoji can be placed randomly anywhere in the sentence. This is called RANDOM-PLACEMENT. The alternative is to place the emojis in a sentence based on a certain distribution and sample the location of placement from that distribution. This method of placement is called LOCATION-SAMPLING. The distribution is found by studying the locations at which different emojis occur in a Twitter feed and fitting the location of their occurrence to a Gaussian distribution. We use the TBv2 train-split to calculate the distribution parameters. We experiment with these two methods for emoji injection for the BERT-base model by injecting tweets in 25\% sentences in the GUM dataset. The models are trained on the transformed dataset and tested on the TBv2 test set. The results are shown in Table \ref{table:emoji}. We find that LOCATION-SAMPLING is significantly superior to the RANDOM-PLACEMENT method of emoji-injection.

\begin{table}
\centering
 \scalebox{0.9}{

    \begin{tabular}{l|l}
    \textbf{Emoji Injection Method} & \textbf{POS Tagging Accuracy} \\
    \hline
    Zero-shot & 79.746 (0.256)\\
    RANDOM-PLACEMENT & 79.103 (0.275)\\
    LOCATION-SAMPLING & 80.125 (0.192)\\ 
    \end{tabular}
    }
    \caption{\label{table:emoji}
    Comparison between random-emoji injection and location-sampling based emoji injection. We find that location-sampling performs significantly better than random placement.
    }
\end{table}

We also experimented with different thresholds for emoji injection. We found that injecting emojis into a larger number of sentences hurts the model performance as shown in  Table \ref{table:emoji2}. Thus, we do emoji injection with a 25\% probability.

\begin{table}
\centering
 \scalebox{0.8}{

    \begin{tabular}{l|l}
    \textbf{Emoji Injection Method} & \textbf{POS Tagging Accuracy} \\
    \hline
    Zero-shot & 79.746 (0.256)\\
    EI (25\%) & 80.123 (0.192)\\
    EI (50\%) & 79.671 (0.331)\\ 
    EI (75\%) & 79.473 (0.389)\\
    \end{tabular}
    }
    \caption{\label{table:emoji2}
    Comparison between different probability thresholds of emoji injection. We find a 25\% probability of adding emojis to a sentence performs optimally.
    }
\end{table}

\subsection{Inverse Lexcical Normalization Ablation}\label{appendix:iln}
Inverse Lexical Normalization (ILN) aims to convert standard English text into its un-normalized versions. This includes converting correct spellings to their noisy versions as used in social media and converting certain texts to corresponding acronyms. Some examples of such a conversion would be converting \textit{you -> u, that - dat, how are you -> hru}.

We do this by using the dataset released by \cite{baldwin2015shared} for lexical normalization. We use the training set as a dictionary and find mappings between the lexically-correct tokens and their noisy usage in social media. When a word in this dictionary is found in the standard English sentence, it is converted into its un-normalized version with a probability of 75\%. The ablation experiments with BERT-base are shown in Table \ref{table:iln}.

\begin{table}
\centering
 \scalebox{0.8}{

    \begin{tabular}{l|l}
    \textbf{ILN Method} & \textbf{POS Tagging Accuracy} \\
    \hline
    Zero-shot & 79.746 (0.256)\\
    ILN (25\%) & 80.329 (0.327)\\
    ILN (50\%) & 80.504 (0.292)\\ 
    ILN (75\%) & 80.668 (0.236)\\
    \end{tabular}
    }
    \caption{\label{table:iln}
    Comparison between different probability thresholds for inverse lexical normalization. This probability threshold is for converting each token in a sentence to its un-normalized version.
    }
\end{table}

\subsection{Injecting User Mentions and Hashtags as PROPN - Ablation}\label{appendix:propn}
User mentions and hashtags are often used as proper nouns (PROPN) as shown in the two examples below : 
\begin{itemize}
    \item \textit{\#FOLLOW us \#CHECKOUT the multi - talented Spanglish Pop Singer Model @USER779 aka Lady Boom Boom URL107}
    \item \textit{Today I went to watch \#Metallica \#themostamazingconcertever}
\end{itemize}

In the first tweet, @USER779 mention is used as a proper noun. In the second example \#Metallica is used as a proper noun followed by another hashtag which refers to a totally different part-of-speech. In this transformation, we convert pre-existing proper nouns in standard English sentences into user mentions or hashtags. In a brief analysis of Twitter feed, we found that user mentions were more common than hashtags. Thus we start by randomly changing proper nouns into user mentions with a probability of 25\% and into hashtags with a probability of 10\%. The ablation experiments with BERT-base model are shown in Table \ref{table:propn}.

\begin{table}
\centering
 \scalebox{0.8}{

    \begin{tabular}{l|l}
    \textbf{PROPN Injection Method} & \textbf{POS Tagging Accuracy} \\
    \hline
    Zero-shot & 79.746 (0.256)\\
    @(25\%), \#(10\%) & 81.604 (0.064)\\
    @(50\%), \#(20\%) & 81.896 (0.061)\\ 
    @(75\%), \#(30\%) & 81.742 (0.055)\\
    \end{tabular}
    }
    \caption{\label{table:propn}
    Comparison between different probability thresholds for converting proper nouns present in standard English sentence to user mentions and hashtags.
    }
\end{table}

\subsection{Injecting Re-Tweets, URLS, User Mentions and Hashtags as X - Ablation}\label{appendix:X}
Re-tweets involving user mentions are separate from when user mentions are used as proper nouns and are classified in the 'X:other' POS category. URLs and some hashtags also fall into this category. Examples of tweets containing these lexical features can be seen in Figure \ref{fig:anatomy}. Injecting these features is simpler than the other lexical features and yet results in the largest improvements. Re-tweets are almost always present at the beginning of a tweet. URLs are almost always present at the end of the tweet. We make a pre-selected list of certain hashtags that fall into the 'X:other' POS tag category and place them randomly in a sentence. We experiment with the relative probability of such injections in Table \ref{table:X}.

\begin{table}
\centering
 \scalebox{0.8}{

    \begin{tabular}{l|l}
    \textbf{X-Injection Method} & \textbf{POS Tagging Accuracy} \\
    \hline
    Zero-shot & 79.746 (0.256)\\
    RT(30\%), URL(60\%), \#(10\%) & 89.334 (0.079)\\
    RT(60\%), URL(60\%), \#(20\%) & 89.253 (0.094)\\ 
    RT(90\%), URL(60\%), \#(30\%) & 89.173 (0.081)\\
    \end{tabular}
    }
    \caption{\label{table:X}
    Comparison between different probability thresholds for injecting re-tweets, URLs and hashtags into a standard English sentence as the POS label `X'.
    }
\end{table}

\subsection{Combining All Lexical Data Transformations}
When we combine all lexical data transformations, we achieve significant boost in performance on the Twitter dataset. When a model trained on the GUM dataset (standard English, source domain) is tested on the Tweebankv2 test set (Twitter dataset, target domain), we see that the model has about 81.52\% accuracy using BERT-large for POS tagging (Table \ref{table:combination_exp}, first row, Unsupervised). When we use all lexical transformations to transform standard English dataset to Twitter like sentences, called GUM-T, we achieve 92.14\% accuracy, and see a significant boost of 10.62\% over the zero-shot performance. This shows us that our simple lexical data transformations give the model a massive boost without training on actual tweets annotated for POS tagging. Our lexical data transformations can be used both for unsupervised domain adaptation and data augmentation, as shown in Table \ref{table:combination_exp}. 

\subsubsection{The `X:other' POS class for Twitter}\label{appendix:class-wise}
The class-wise F1 score improvements in BERT-large for unsupervised domain adaptation are shown in Table \ref{table:classwise}. We see significant improvements for all POS classes. The improvement is massive for the `X' POS class because this class works very differently in standard English and tweets. Tweets contain a lot of hashtags, URLs, and re-tweets, which is completely different from standard English. Thus, the `X' POS class is the biggest lexical differentiator between standard English and how people communicate on Twitter. This is also why the performance of a POS tagger trained on standard English dataset performed abysmally, with and F1 score of 0.01. 

\begin{table}
\centering
 \scalebox{0.8}{

    \begin{tabular}{l|l|l|l}
    \textbf{POS class} & \textbf{Zero-Shot F1} & \textbf{Transformed F1} & \textbf{Tokens} \\
    \hline
    NOUN & 0.85 & 0.87 & 2669 \\
    NUM & 0.80 & 0.92 & 304\\
    PROPN & 0.63 & 0.97 & 1716\\
    SYM & 0.53 & 0.79 & 209\\
    VERN & 0.87 & 0.93 & 1985\\
    X & 0.01 & 0.94 & 2056\\ 
    \end{tabular}
    }
    \caption{\label{table:classwise}
    Class-wise F1 improvement for unsupervised domain adaptation for BERT-large model for selected classes. Zero-shot F1 shows the class-wise F1 scores before applying lexical data transformations. Transformed F1 shows the class-wise F1 score for unsupervised domain adaptation of BERT-large model after application of lexical data transformations. 
    }
\end{table}

\subsection{Lexicalally Transformed Sentences}
\begin{figure}
 \centering
 \scalebox{0.35}{
    \centering
     \includegraphics{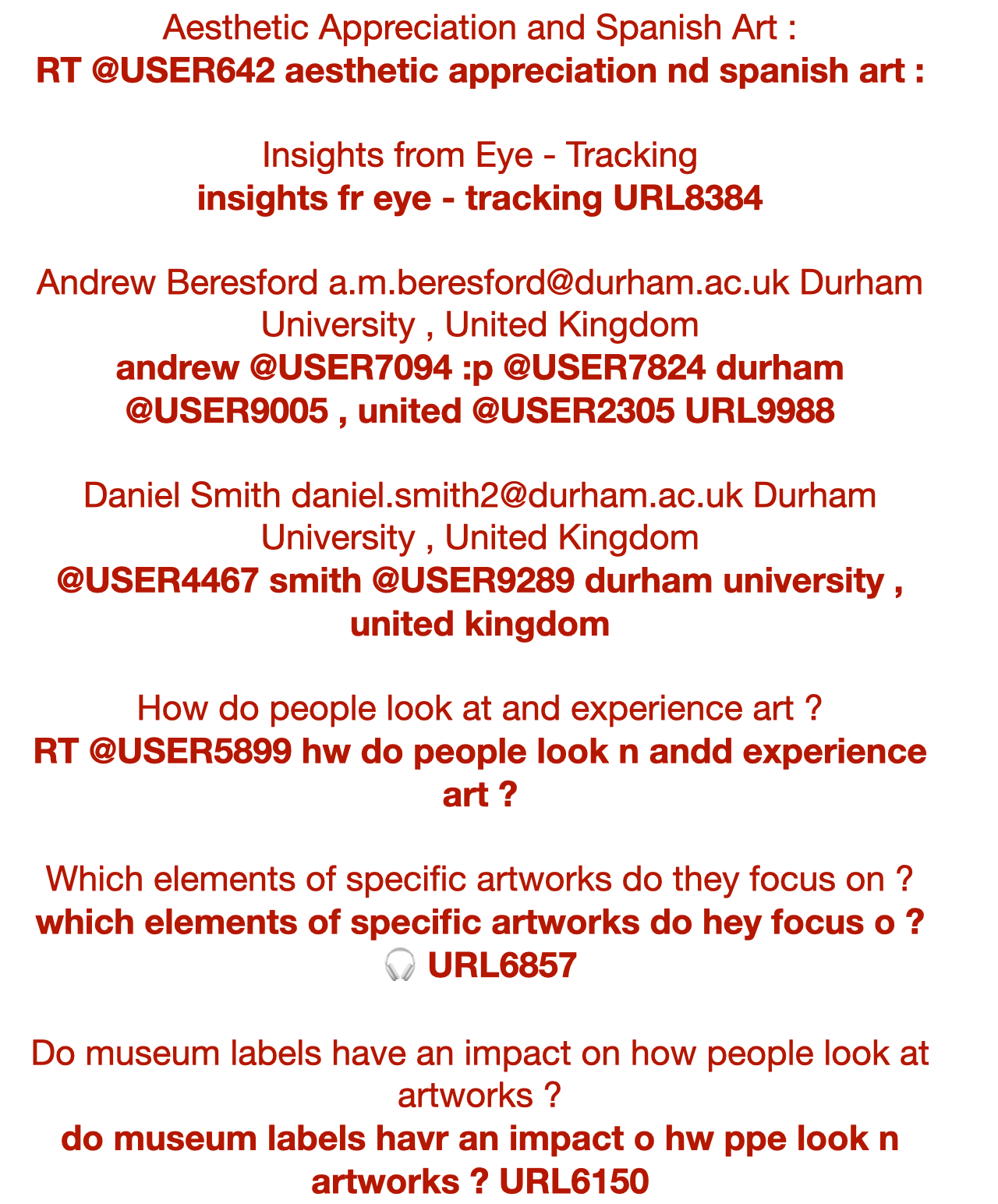}}
     \caption{Figure shows various examples of lexicalally transformed standard English sentences. The sentence in the bold font corresponds to the lexicalally transformed sentence for the original standard English sentence written directly above it regular font.}
     \label{fig:moreexample}
\end{figure}

Some examples of the lexicalally transformed sentences from standard English to tweets are shown in Figure \ref{fig:moreexample}. The examples show different features including emojis, user mentions, re-tweets, URLs and lexically incorrect tokens. 

\subsection{Average Runtimes, Hyperparameters and Hardware}
All experiments were performed on a single Tesla T4 GPU with 16GB GPU memory in a system with 16GB RAM. The run-time for base models per epoch was approximately 2 minutes for the Tweebank train-split and 6 minutes for the GUM train-split. For large models, the time taken per epoch was approximately 6 minutes for Tweebank train-split and 18 minutes for GUM train-split. The best performance and best dev-accuracy were chosen. We kept a batch size of 32, a learning rate of 1e-5 and maximum sequence length of 256. All models are trained for 25 epochs. We run each configuration 5 times and report the mean scores and standard deviation.

\end{document}